\def\year{2022}\relax
\documentclass[letterpaper]{article} 
\usepackage{aaai22}  
\usepackage{times}  
\usepackage{helvet}  
\usepackage{courier}  
\usepackage[hyphens]{url}  
\usepackage{graphicx} 
\usepackage{natbib}  
\usepackage{caption} 
\DeclareCaptionStyle{ruled}{labelfont=normalfont,labelsep=colon,strut=off} 
\usepackage[ruled,vlined]{algorithm2e}
\usepackage{multirow}
\usepackage{multicol}
\usepackage{makecell}
\usepackage{threeparttable}
\usepackage{bm}
\usepackage{amsmath}
\usepackage{amssymb}
\usepackage[switch]{lineno}
\usepackage{color}

\title{Multiple Domain Experts Collaborative Learning: \\ Multi-Source Domain Generalization For Peron Re-Identification}

\author {
    Shijie Yu,\textsuperscript{\rm 1,2}\thanks{This work was done when Shijie Yu was an intern at SenseTime (sj.yu@siat.ac.cn).}
    Feng Zhu,\textsuperscript{\rm 3}
    Dapeng Chen,\textsuperscript{\rm 3}
    Rui Zhao,\textsuperscript{\rm 3,5}
    Haobin Chen,\textsuperscript{\rm 1,2} \\
    Jinguo Zhu,\textsuperscript{\rm 3}
    Shixiang Tang,\textsuperscript{\rm 3}
    Yu Qiao,\textsuperscript{\rm 1,4}\thanks{Corresponding author (yu.qiao@siat.ac.cn).}
}
\affiliations{
    \textsuperscript{\rm 1}ShenZhen Key Lab of Computer Vision and Pattern Recognition, \\Shenzhen Institutes of Advanced Technology, Chinese Academy of Sciences \\
    \textsuperscript{\rm 2}University of Chinese Academy of Sciences, China~~~~
    \textsuperscript{\rm 3}SenseTime Group Limited~~~~
    \textsuperscript{\rm 4}Shanghai AI Lab, Shanghai, China \\
    \textsuperscript{\rm 5}Qing Yuan Research Institute, Shanghai Jiao Tong University, Shanghai, China 
}

\usepackage{bibentry}

\begin{document}
\maketitle

\begin{abstract}
    Recent years have witnessed significant progress in person re-identification (ReID). However, current ReID approaches still suffer from considerable performance degradation when unseen testing domains exhibit different characteristics from the source training ones, known as the domain generalization problem. Given multiple source training domains, previous Domain Generalizable ReID (DG-ReID) methods usually learn all domains together using a shared network, which can't learn sufficient knowledge from each domain. In this paper, we propose a novel \emph{\textbf{M}ultiple Domain \textbf{E}xperts \textbf{C}ollaborative \textbf{L}earning (MECL)} framework for better exploiting all training domains, which benefits from the proposed Domain-Domain Collaborative Learning (DDCL) and Universal-Domain Collaborative Learning (UDCL). DDCL utilizes domain-specific experts for fully exploiting each domain, and prevents experts from over-fitting the corresponding domain using a meta-learning strategy. In UDCL, a universal expert supervises the learning of domain experts and continuously gathers knowledge from all domain experts.
    Note, only the universal expert will be used for inference. Extensive experiments on DG-ReID benchmarks demonstrate the effectiveness of DDCL and UDCL, and show that the whole MECL framework significantly outperforms state-of-the-arts. Experimental results on DG-classification benchmarks also reveal the great potential of applying MECL to other DG tasks. Code will be released.

\end{abstract}

\section{Introduction}
Person re-identification (ReID) which aims to associate the corresponding person across non-overlapped cameras given query person images or videos, has attracted more and more attention due to its promising application in public security and smart city. Recently, person ReID methods \cite{reid_chen2019abd, reid_hou2019interaction, reid_liu2020unity} based on deep learning have achieved significant performance improvement. However, an assumption in their settings is that the training set and testing set are collected from the same domain, which limits their practical applications because the domains vary with the background, illumination and so on in the real-world scenarios, leading to drastic performance degradation of ReID models. 
Unsupervised domain adaptation (UDA) ReID methods \cite{uda_fu2019self, uda_ge2020mutual, uda_kumar2020unsupervised, uda_song2020unsupervised} tackle the domain shift problem in a domain adaptation manner, that is to adapt a trained model to the target domain based on unlabeled target-domain training data, but can not guarantee the performance on \textbf{unseen} target domains.



\begin{figure}[t]
    \centering
    \includegraphics[width=1.0\linewidth]{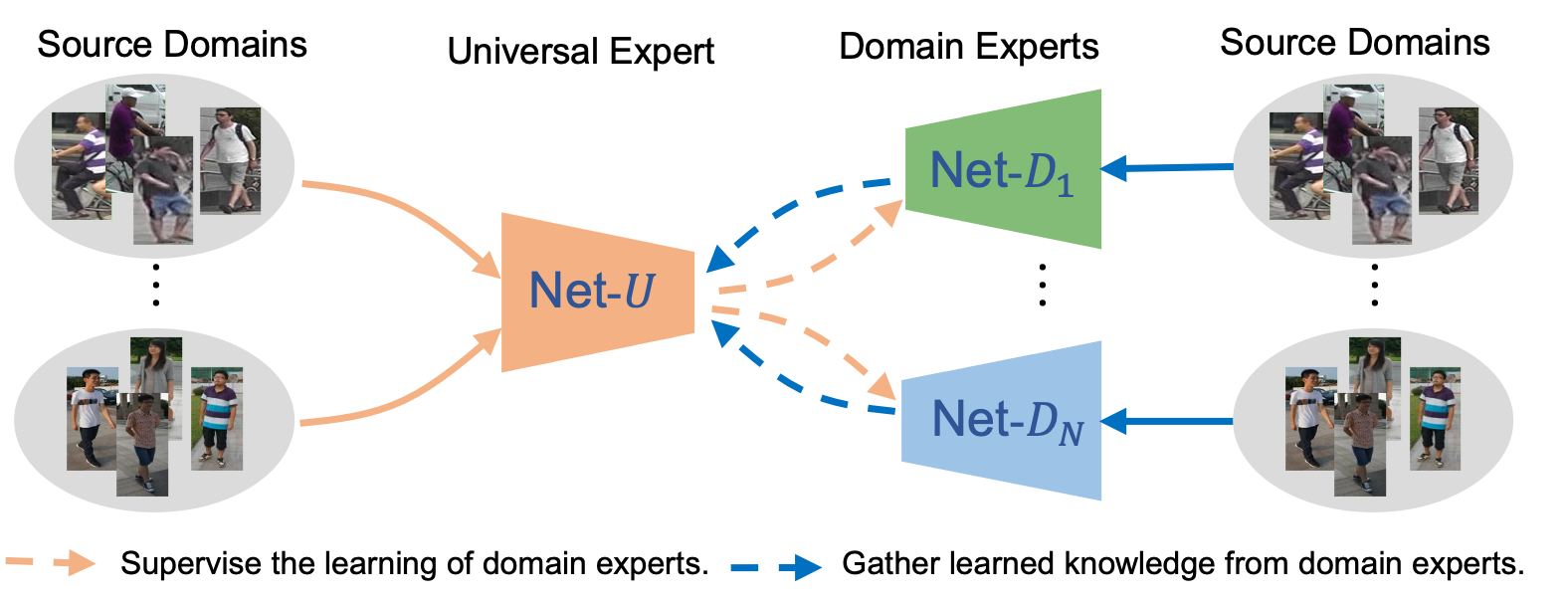}
    \caption{The overall framework of our MECL. Each domain is associated with a specific expert. Domain experts and the universal expert are jointly learned through domain-domain and universal-domain collaborative learning. The universal expert is used for inference.}
    \label{fig:frameworks}
\end{figure}

Compared with UDA, domain generalization (DG) \cite{DG_adver_li2018domain, DG_muandet2013domain} is more challenging but practical because it doesn't require any prior knowledge about the target domain during training, that is, target domains are unknown to the models. DG methods aim to improve the model's generalization capability across domains so that they can be applied to any unseen domain once trained, i.e., ``train once, run everywhere". Most previous DG methods \cite{DG_matsuura2020domain, DG_meta_li2018learning, DG_muandet2013domain} reach a consensus that the data from different domains share the same label space. However, there is usually no ID overlap between source domains and target domains in ReID tasks, making domain generalization learning for ReID (DG-ReID) more challenging. Following \citet{zhao2020learning},  we focus on a more practical setting of DG-ReID, referred as multi-source DG-ReID, where the datasets for training are collected from multiple domains.



 Existing works \cite{metaBIN, zhao2020learning} on the multi-source DG-ReID task usually share the same feature extractor among multiple domains. We argue that it can't exploit each domain sufficiently because the optimization directions of different domains may have dramatic differences which will affect the learning of the single network. To alleviate this problem, we propose a novel model-agnostic learning framework (as shown in Fig. \ref{fig:frameworks}), named \textit{\textbf{M}ultiple Domain \textbf{E}xperts \textbf{C}ollaborative \textbf{L}earning (MECL)}, which jointly trains multiple domain-specific experts and one universal expert in a collaborative learning manner. Compared to other single model based methods, \emph{MECL requires no additional computation cost in testing as only the universal expert will be used for inference}.
 
 In greater detail, two collaborative learning algorithms work together organically in MECL: (1) \emph{Domain-Domain Collaborative Learning} (DDCL). Each domain expert associated with a specific domain concentrates on exploiting the corresponding domain sufficiently. Meanwhile, to avoid over-fitting one specific domain, domain experts will communicate with others to regulate its learning procedure during training using a meta-learning strategy. Vividly speaking, a good expert should not only be good at his major but also know something about other areas. (2) \emph{Universal-Domain Collaborative Learning} (UDCL). The universal expert supervises the learning of domain experts and continuously gathers learned knowledge from all domain experts.
 Specifically, at the beginning of each training iteration, the universal expert will review the duplicated mini-batch data from all source domains and then provide supervision to the domain experts through the alignment loss and uniformity loss. At the end of the iteration, it will gather the learned knowledge from all domain experts to update its parameters by exponential moving average (EMA). DDCL and UDCL complement each other, and can significantly improve the generalization capability of learned models when applied together in our MECL framework.
 
 

In summary, the main contributions of this paper are three-fold: (1) We propose a novel model-agnostic learning framework called \textbf{M}ultiple Domain \textbf{E}xperts \textbf{C}ollaborative \textbf{L}earning (MECL) for multi-source Domain Generalizable person ReID (DG-ReID), in which Domain-Domain Collaborative Learning (DDCL) and Universal-Domain Collaborative Learning (UDCL) organically work together to improve the model's generalization capability across domains. (2) We establish a simple but rather strong multi-source DG-ReID baseline method named Multi-Domain Equality (MDE) which outperforms the conventional baseline by a large margin. This strong baseline method will facilitate future works in this area. (3) We perform extensive experiments on both DG-ReID and DG-classification benchmarks, not only demonstrate the effectiveness of our MECL framework on improving the model's generalization capability for person ReID, but also reveal the great potential of applying MECL to other DG tasks.


\section{Related Work}
\noindent \textbf{Person ReID.} 
Person ReID \cite{reid_chen2019abd, reid_li2014deepreid, reid_sun2018beyond, reid_park2020relation, reid_luo2019bag, reid_li2018harmonious} based on deep learning has made remarkable progress recently. However, these methods mainly focus on learning discriminative intra-domain person features, that is, training and evaluating on the same domain, ignoring the model's generalization capability to unseen domains. As shown by \citet{reid_luo2019bag}, the model trained on Market1501 \cite{market1501} dataset suffers from dramatic performance degradation when it is tested on DukeMTMC-reID \cite{duke1}, which heavily impedes the practical applications of ReID systems. Recently, Unsupervised domain adaptation (UDA) methods \cite{uda_kumar2020unsupervised, uda_zhai2020ad, uda_song2020unsupervised} are proposed to adapt ReID models from a labeled source domain to an unlabeled target domain. This adaptation paradigm requires amounts of unlabeled target-domain training data, thus can not guarantee the performance on unseen target domains.

\noindent \textbf{Domain Generalization (DG).}
DG is more challenging but practical than UDA, because it doesn't require any data of target domains during training. Recent works on this topic mainly concentrate on (1) learning domain-invariant features by minimizing the inter-domain discrepancy of the same identity \cite{akuzawa2019domain,xiao2021variational, DG_adver_li2018domain}, or (2) optimizing the network using meta-learning strategy to improve the generalization capability \cite{DG_meta_li2018learning, dou2019domain}. These DG methods are usually developed on classification tasks, where source and target domains share the same label space. However, for DG-ReID tasks, there are few overlapped identities across domains, which makes DG-ReID more challenging.


\noindent \textbf{Domain Generalization for Person ReID.}
Conventional DG methods developed on classification can not be directly applied to DG-ReID as different domain doesn't share label space. There are three main categories of DG-ReID methods. (1) Normalization-based methods \cite{SNR, metaBIN, jia2019frustratingly}. These methods mainly utilize batch normalization (BN) \cite{ioffe2015batch} and instance normalization (IN) \cite{ulyanov2016instance} to filter out the identity-irrelevant information.
\citet{SNR} proposed the style normalization and restitution (SNR) module based on IN to further disentangle the identity-relevant features and identity-irrelevant features. (2) Adversarial learning based methods \cite{lin2020multi, tamura2019augmented}. 
\citet{lin2020multi} employed an adversarial auto-encoder module and a discriminator to guide feature extractor to extract domain-invariant features across domains. (3) Meta-learning based methods \cite{metaBIN, DG_meta_li2018learning, song2019generalizable}.
\citet{zhao2020learning} proposed M$^3$L approach which employs meta-learning to train the whole feature extractor network. \citet{metaBIN} proposed the MetaBIN that not only employs normalization layers, but also uses meta-learning to learn the balance weight of BIN layers. 

The above methods usually share a single network among multiple source domains. Differently, our MECL framework assigns a network (domain expert) to each source domain to exploit the source domain sufficiently, and jointly train multiple domain experts and a universal expert through domain-domain and universal-domain collaborative learning. Besides, MECL is a general DG framework that can be applied to both DG-ReID and DG-classification tasks.




\noindent \textbf{Collaborative Learning.} Our collaborative learning is also inspired by some similar works from the semi-supervised learning \cite{mean-teacher, co-training}, the knowledge distillation \cite{hinton2015distilling, tian2019contrastive}, the self-supervised learning \cite{grill2020bootstrap, moco, siamese}, \emph{etc}. In ReID, \citet{uda_ge2020mutual} proposed the mutual mean-teaching framework and \citet{zhai2020multiple} proposed the multiple experts brainstorming method for the UDA task.

\section{Methodology}
\subsection{Problem Definition}
At the beginning, we formally give the definition of the multi-source DG-ReID problem. Assume that we have access to $N$ source domains, i.e., $N$ person ReID datasets denoted as $\mathcal{D}^s = \{\mathcal{D}^s_n\}_{n=1}^N$ for training, and $M$ target domains denoted as $\mathcal{D}^t = \{\mathcal{D}^t_m\}_{m=1}^M$ for testing. Note that there is no overlap between source and target domains, which means $\mathcal{D}^s \cap \mathcal{D}^t = \emptyset$. The $k$-th source domain $\mathcal{D}^s_k \in \mathcal{D}^s$ with $P^k$ images is denoted as $\{(\mathcal{I}_i^k, y_i^k)\}^{P^k}_{i=1}$ where $\mathcal{I}_i^k$ is the $i$-th image and $y^k_i$ is the corresponding identity label from the label space $\mathcal{Y}^k$. Different with the DG-classification problem, the source domains in the multi-source DG-ReID don't share the label space, i.e., $\bigcap_{n=1}^N\mathcal{Y}^n = \emptyset$. The goal of the multi-source DG-ReID is to fully exploit the $N$ source domains to train a more generalizable model which is expected to have a better performance on the $M$ target domains.


\begin{figure*}[t]
    \centering
    \includegraphics[width=1.0\textwidth]{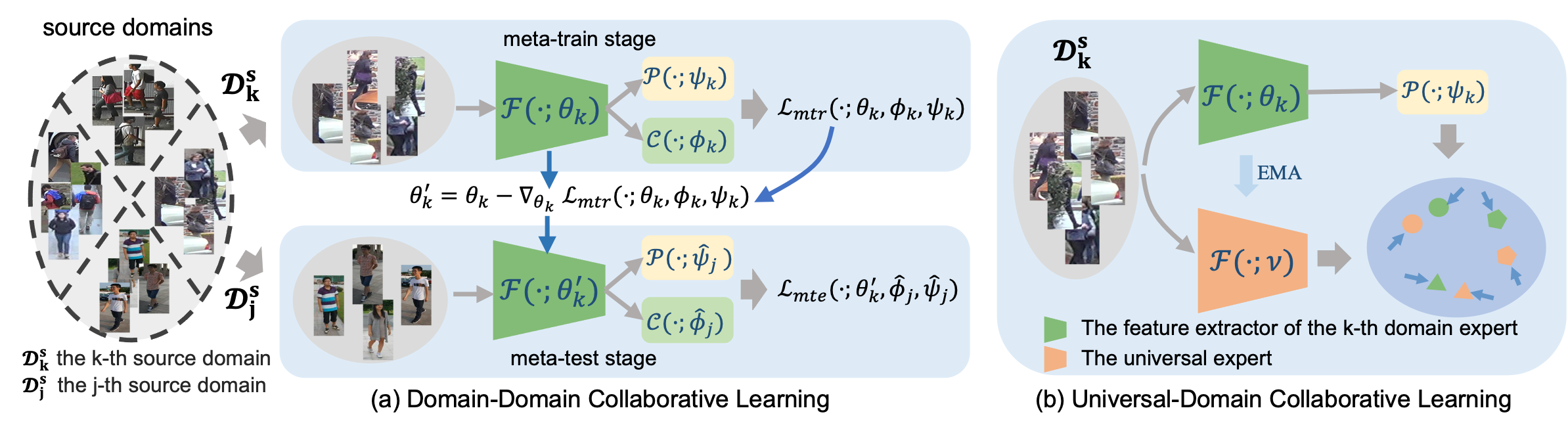}
    \caption{The learning process of the $k$-th domain expert. (a): \textit{Domain-Domain Collaborative Learning}. The collaboration among domain experts is mainly reflected in the meta-test stage, where the feature extractor, classifier and projector will be combined dynamically in terms of the meta-test domain.
    (b): \textit{Universal-Domain Collaborative Learning}. The universal expert supervises the learning of the domain experts by the alignment loss and the uniformity loss. 
    Besides, the universal expert gathers the learned knowledge from the domain experts in the EMA manner. Points of the same shape denote feature embeddings from the same image, while the color indicates corresponding feature extractor.}
    
    \label{fig:learning}
\end{figure*}

\subsection{Baseline Methods}
We introduce the two baseline methods used in this paper in detail, including the traditional DAG baseline and our proposed stronger MDE baseline.

\noindent \textbf{Domain Aggregation (DAG) Baseline.} The DAG baseline is the most commonly used baseline approach in multi-source DG-ReID \cite{metaBIN, lin2020multi, zhao2020learning}. In particular, it firstly merges all the source datasets into a single dataset $\mathcal{D}^s_{agg}$ without regard to which domain they are from, where $\mathcal{D}^s_{agg} = \bigcup_{n=1}^N\mathcal{D}^s_n$. Then, the model will be trained on $\mathcal{D}^s_{agg}$ in a standard ReID training procedure \cite{reid_luo2019bag}. There are at least three reasons that cause this baseline is not sufficient: (1) The label space of $D^s_{agg}$ is relatively larger, which makes the training of the classifier more difficult. (2) The source datasets are usually not balanced, and the large-scale datasets will be dominated in $D^s_{agg}$. (3) The domain discrepancy is totally ignored in training. We introduced another stronger baseline setting inspired by multi-task learning as follows.

\noindent \textbf{Multi-Domain Equality (MDE) Baseline.} The MDE baseline has addressed the aforementioned three deficiencies of the DAG baseline from the view of multi-task learning. In the settings of MDE, each source domain has its own classifier and shares the feature extractor network with others. During training, all domains are treated equally. Specifically, in each iteration, we will sample a mini-batch with $B$ images from each domain for training, denoted as $\{(\bm{\mathcal{I}}^n,\bm{y}^n)\}_{n=1}^N$, and the loss is formulated as follows:
\begin{equation}
    \mathcal{L}_{mde} = \frac{1}{N}\sum_{n=1}^N (\mathcal{L}_{id}(\bm{\mathcal{I}}^n,\bm{y}^n)+\lambda\mathcal{L}_{tri}(\bm{\mathcal{I}}^n,\bm{y}^n)).
    \label{eq:mde}
\end{equation}
The $\mathcal{L}_{id}$ and $\mathcal{L}_{tri}$ balanced by $\lambda$ are the commonly used softmax classification loss (cross entropy loss) and triplet loss, which are in form of
\begin{equation}
    \mathcal{L}_{id} = \frac{1}{B}\sum\limits_{i=1}^B\mathcal{L}_{ce}(\mathcal{C}(\mathcal{F}(\bm{\mathcal{I}}_i^n;\bm{\theta});\bm{\phi}_n), \bm{y}_i^n),
    \label{loss:ce}
\end{equation}
\begin{equation}
\begin{aligned}
    \mathcal{L}_{tri} = \frac{1}{B}\sum\limits_{i=1}^B\max(&||(\mathcal{F}(\bm{\mathcal{I}}_i^n;\bm{\theta})-\mathcal{F}(\bm{\mathcal{I}}_{i,+}^n;\bm{\theta}))||+ m \\-  &||\mathcal{F}(\bm{\mathcal{I}}_i^n;\bm{\theta}) - \mathcal{F}(\bm{\mathcal{I}}_{i,-}^n;\bm{\theta})||, 0),
\end{aligned}
\label{loss:tri}
\end{equation}
where $\mathcal{F}(\cdot;\bm{\theta})$ and $\mathcal{C}(\cdot;\bm{\phi}_n)$ denote the shared feature extractor and the $n$-th domain classifier, $\bm{\mathcal{I}}_{i,+}^n$ and $\bm{\mathcal{I}}_{i,-}^n$ indicate the farthest positive and nearest negative sample of $\bm{\mathcal{I}}_{i}^n$, and $m$ fixed to 0.3 is the triplet distance margin. Extensive experiments have shown that the MDE baseline has a better performance on unseen domains than the DAG baseline.

\subsection{Multiple Domain Experts Collaborative Learning}
In this subsection, we dive into the proposed \textit{Multiple Domain Experts Collaborative Learning (MECL)} framework. The network architecture and the collaborative learning algorithm will be detailedly explained.

\noindent \textbf{Overall Framework.}
The overall framework of MECL is illustrated in Fig. \ref{fig:frameworks}, which mainly consists of $N$ domain experts and one universal expert. During training, the domain experts and the universal expert interact with each other through the proposed collaborative learning approach, which can be further divided into domain-domain collaborative learning (DDCL) and universal-domain collaborative learning (UDCL). The detailed end-to-end algorithm of MECL is shown in Algorithm. \ref{Alg:MECL}. 

\noindent \textbf{Network Architecture.}
Each domain expert is composed of three components, a feature extractor, a classifier and a projector. For simplicity, we formulate the $N$ domain experts as $\{\mathcal{F}(\cdot;\bm{\theta}_n)\}_{n=1}^{N}$, $ \{\mathcal{C}(\cdot;\bm{\phi}_n)\}_{n=1}^{N}$ and $\{\mathcal{P}(\cdot;\bm{\psi}_n)\}_{n=1}^{N}$ in terms of the component categories, where $\bm{\theta}_n, \bm{\phi}_n, \bm{\psi}_n$ denote the corresponding model parameters of the $n$-th domain expert. The universal expert only consists of a feature extractor parameterized by $\bm{\nu}$, denoted as $\mathcal{F}(\cdot;\bm{\nu})$. Note that, the feature extractors of the domain experts and the universal expert use the same type of backbones. In addition, \emph{the universal expert is employed to extract features in the inference stage}.

\begin{algorithm}[ht!]
\KwIn{$N$ source domains: $\{\mathcal{D}^s_n\}_{n=1}^N$;
$N$ domain experts, including the feature extractors: $\{\mathcal{F}(\cdot ;\bm{\theta}_n)\}_{n=1}^N$, the classifiers: $\{\mathcal{C}(\cdot ;\bm{\phi}_n)\}_{n=1}^N$ and the projectors: $\{\mathcal{P}(\cdot ;\bm{\psi}_n)\}_{n=1}^N$; One universal expert: $\mathcal{F}(\cdot;\bm{\nu})$;
Maximum training iteration $T^*$; Learning rate $\beta$; Step size $\alpha$; Ensembling momentum $\epsilon$;
}

\KwOut{The universal expert $\mathcal{F}(\cdot;\bm{\nu}^*)$.}
\textbf{Initialization}: $\bm{\nu}^{(0)} = \bm{\theta}_1^{(0)} = \cdot\cdot\cdot= \bm{\theta}_N^{(0)}$;\\
~~~~~~~~~~~~~~~~~~~~~~~~$\bm{\psi}_1^{(0)} =\cdot\cdot\cdot= \bm{\psi}_N^{(0)}$. \\
\For{$T \gets 1 $ \KwTo $T^*$}{
     Sample $N$ mini-batches from $N$ source domains, denoted as $\{(\bm{\mathcal{I}}^n,\bm{y}^n)\}_{n=1}^N$; \\
     \For{$k\gets1$ \KwTo $N$}{
     \textbf{Meta-Train:} \\
     Select $(\bm{\mathcal{I}}^k,\bm{y}^k) \in \{(\bm{\mathcal{I}}^n,\bm{y}^n)\}_{n=1}^N$ as the meta-train data;\\
     Compute the meta-train loss:  $\mathcal{L}_{mtr}(\bm{\mathcal{I}}^k,\bm{y}^k;\bm{\theta}_k, \bm{\phi}_k, \bm{\psi}_k)$; \\
     Compute the adapted parameter: $\bm{\theta}_k^{'} = \bm{\theta}_k - \alpha \nabla_{\bm{\theta}_k} \mathcal{L}_{mtr}(\bm{\mathcal{I}}^k,\bm{y}^k;\bm{\theta}_k, \bm{\phi}_k, \bm{\psi}_k)$; \\
     \textbf{Meta-Test:} \\
     Randomly select the meta-test data $(\bm{\mathcal{I}}^j,\bm{y}^j)$ from $\{(\bm{\mathcal{I}}^n,\bm{y}^n)\}_{n=1}^N$ where $j \neq k$;\\
     Compute the meta-test loss using $\bm{\theta}_k^{'}$:  $\mathcal{L}_{mte}(\bm{\mathcal{I}}^j,\bm{y}^j;\bm{\theta}_k^{'}, \hat{\bm{\phi}}_j, \hat{\bm{\psi}}_j)$ \\
     \textbf{Optimization of the $k$-th domain expert:} \\
     Compute the uniformity loss $\mathcal{L}_{unif}$; \\
     Compute the overall loss $\mathcal{L}_{total}$ ; \\
     Update the model parameters of the $k$-th domain expert: \\
     ~~$\bm{\theta}_k^{(T)} = \bm{\theta}_k^{(T-1)} - \beta\nabla_{\bm{\theta}_k}\mathcal{L}_{total}$; \\
     ~~$\bm{\phi}_k^{(T)} = \bm{\phi}_k^{(T-1)} - \beta\nabla_{\bm{\phi}_k}\mathcal{L}_{total}$;\\
     ~~$\bm{\psi}_k^{(T)} = \bm{\psi}_k^{(T-1)} - \beta\nabla_{\bm{\psi}_k}\mathcal{L}_{total}$.
    }
    \textbf{Optimization of the universal expert:} \\
    Update the model parameters of the universal expert using exponential moving average:
    ~~~~$\bm{\nu}^{(T)} = \epsilon~\bm{\nu}^{(T-1)} + (1-\epsilon) \frac{1}{N}\sum_{n=1}^N\bm{\theta}_n^{(T)}$.
}
At the end of the training: $\mathcal{F}(\cdot;\bm{\nu}^*) = \mathcal{F}(\cdot;\bm{\nu}^{(T^*)})$.
\caption{{\bf Multiple Domain Experts Collaborative Learning (MECL)} 
\label{Alg:MECL}}
\end{algorithm}


\noindent \textbf{Domain-Domain Collaborative Learning.} 
To avoid overfitting the specific domains, the domain experts should communicate with others periodically to regulate their learning process. Following \citet{DG_meta_li2018learning,zhao2020learning,song2019generalizable}, we apply the model-agnostic meta-learning (MAML) \cite{finn2017model} to the training of each domain expert, because it can not only further improve the generalization capability of models \cite{DG_meta_li2018learning}, but also strengthen the interaction among domain experts by dynamically combining the three components of \emph{the feature extractors, classifiers and projectors} in the meta-test stage as shown in Fig. \ref{fig:learning}-a.

At the beginning of each training iteration, we will randomly sample a mini-batch with $B$ image-label pairs from each domain, denoted as $\{(\bm{\mathcal{I}}^n,\bm{y}^n)\}_{n=1}^N$, where $(\bm{\mathcal{I}}^n,\bm{y}^n)$ comes from $\mathcal{D}^s_n$. Take the training of the $k$-th domain expert for example. The mini-batch $(\bm{\mathcal{I}}^k,\bm{y}^k)$ is treated as meta-train samples while the meta-test samples  $(\bm{\mathcal{I}}^j,\bm{y}^j)$ are randomly selected from the left mini-batches, that is,  $(\bm{\mathcal{I}}^j,\bm{y}^j)\in\{(\bm{\mathcal{I}}^n,\bm{y}^n)\}_{n=1}^N$ where $j \neq  k$. 

In the meta-train stage, the meta-train loss with respect to $(\bm{\mathcal{I}}^k, \bm{y}^k)$ is denoted as $\mathcal{L}_{mtr}(\bm{\mathcal{I}}^k,\bm{y}^k;\bm{\theta}_k, \bm{\phi}_k, \bm{\psi}_k)$, where $\bm{\theta}_k, \bm{\phi}_k, \bm{\psi}_k$ are the model parameters of the $k$-th domain expert. Moreover, $\mathcal{L}_{mtr}$ is the combination of Eq. \ref{loss:ce}, \ref{loss:tri} and \ref{loss:align}, i.e., $\mathcal{L}_{mtr} = \mathcal{L}_{id} + \mathcal{L}_{tri} + \mathcal{L}_{align}$. Then, we compute the adapted parameters of $\bm{\theta}_k$ by 
\begin{equation}
    \bm{\theta}_k^{'} = \bm{\theta}_k - \alpha \nabla_{\bm{\theta}_k} \mathcal{L}_{mtr}(\bm{\mathcal{I}}^k,\bm{y}^k;\bm{\theta}_k, \bm{\phi}_k, \bm{\psi}_k),
    \label{eq:adapt}
\end{equation}
where $\alpha$ is the step size which is fixed to 0.1 here. Note that, only $\bm{\theta}_k$ needs to be meta-learned.

In the meta-test stage, the meta-test loss with respect to $(\bm{\mathcal{I}}^j, \bm{y}^j)$ should be calculated under the condition of $\bm{\theta}_k^{'}$, denoted as
$\mathcal{L}_{mte}(\bm{\mathcal{I}}^j,\bm{y}^j;\bm{\theta}_k^{'}, \hat{\bm{\phi}}_j, \hat{\bm{\psi}}_j)$,
where $\hat{\bm{\phi}}_j, \hat{\bm{\psi}}_j$  belong to the $j$-th domain expert, and $\hat{\cdot}$ denotes that the parameters will not be optimized here. Note, $\mathcal{L}_{mte}$ and $\mathcal{L}_{mtr}$ have the same form with different inputs and parameters.

Finally, we combine $\mathcal{L}_{mtr}$ and $\mathcal{L}_{mte}$ to optimize $\bm{\theta}_k, \bm{\phi}_k, \bm{\psi}_k$, respectively, \emph{i.e.}
\begin{equation}
    \mathop{\arg\min}\limits_{\bm{\theta}_k, \bm{\phi}_k, \bm{\psi}_k}\frac{1}{2}(\mathcal{L}_{mtr} + \mathcal{L}_{mte}).
    \label{eq: meta}
\end{equation}

\noindent \textbf{Universal-Domain Collaborative Learning.} 
During the training stage of MECL, the universal expert takes responsibility for providing supervision to the domain experts and periodically gathering what they have learned to improve itself. Each domain expert and the universal expert will learn mutually as illustrated in Fig. \ref{fig:learning}-b. 

At first, the universal expert will review the $N$ mini-batch data, i.e., project the images into the feature vectors by $\mathcal{F}(\cdot;\bm{\nu})$. Then, $\mathcal{F}(\bm{\mathcal{I}}^k;\bm{\nu})$ will be used to supervise the $k$-th domain expert  using the alignment loss in terms of 
\begin{equation}
    \mathcal{L}_{align} = \frac{1}{B}\sum_{i=1}^B||\mathcal{P}(\mathcal{F}(\bm{\mathcal{I}}^k_i;\bm{\theta}_k);\bm{\psi}_k) - \mathcal{F}(\bm{\mathcal{I}}^k_i;\bm{\nu})||,
    \label{loss:align}
\end{equation}
where $||\cdot||$ denotes the Euclidean distance between two feature vectors. The projector $\mathcal{P}(\cdot;\bm{\psi}_k)$ \cite{grill2020bootstrap} attempts to bridge the gap between the universal expert and the $k$-th domain expert to make the optimization easier.

The alignment loss provides the supervision from the perspective of pushing the positive samples (two types of features of the same image). Inspired by \citet{wang2020understanding}, we employ the uniformity loss which fully exploits the negative samples to encourage the feature distribution more uniform among domains. For the training of $k$-th domain expert,  it is defined as follows: 
\begin{equation}
    \mathcal{L}_{unif} = \frac{1}{B}\sum_{i=1}^B\log\frac{1}{N}\sum_{n=1}^N(\frac{1}{Q}\sum^Q_{q=1}e^{-2||\bm{f}^k_{i} - \bm{\bar{f}}^n_{q}||}),
    \label{loss:unif}
\end{equation}
where $\bm{f}_i^k = \mathcal{F}(\bm{\mathcal{I}}_i^k;\bm{\theta}_k)$, $\bm{\bar{f}}_{q}^n = \mathcal{F}(\bm{\bar{\mathcal{I}}}_{q}^n;\bm{\nu})$, $\bm{\bar{\mathcal{I}}}_{q}^n$ is the negative sample of $\bm{\mathcal{I}}_{i}^k$ and $Q$ is the number of the negative samples in the $n$-th domain. Actually, minimizing $\mathcal{L}_{unif}$ is equal to maximize the distance of a sample of $D^s_k$ to its negative samples which are sampled from $\{D^s_n\}_{n=1}^N$. Intuitively, a sample is closer to its negative samples from the same domain than those from others, but the ideal situation is that a sample should be far away from the negative samples no matter what the domain they are from, $\mathcal{L}_{unif}$ is therefore used to make each sample keep away from the negative samples from any domain.

After one iteration of all domain experts, the universal expert will gather the learned knowledge from each domain expert to improve itself. We update the parameters of the universal expert in the manner of the exponential moving average (EMA) \cite{mean-teacher}, which is defined as follows:
\begin{equation}
    \bm{\nu}^{(T)} = \epsilon~\bm{\nu}^{(T-1)} + (1-\epsilon) \frac{1}{N}\sum_{n=1}^N\bm{\theta}_n^{(T)},
    \label{eq:ema}
\end{equation}
where $\bm{\nu}^{(T-1)}$ denotes the parameters of the universal expert in the previous iteration $(T-1)$, $\bm{\theta}_n^{(T)}$ is the feature extractor parameters of the $n$-th domain expert in the current iteration $(T)$, and $\epsilon$ is the ensembling momentum usually set to 0.999. The initialization of these parameters are $\bm{\nu}^{(0)} = \bm{\theta}^{(0)}_1 = \cdot\cdot\cdot = \bm{\theta}_N^{(0)}$.

Totally, the overall loss function are in the form of 
\begin{equation}
\mathcal{L}_{total} = \frac{1}{2}(\mathcal{L}_{mtr}+ \mathcal{L}_{mte}) + \gamma \mathcal{L}_{unif},
\label{loss:total}
\end{equation}
where $\gamma = 0.1$ is to balance the influence of  $\mathcal{L}_{unif}$.

\begin{table*}
    \centering
    \caption{Ablation studies of MECL. `DDCL' is the domain-domain collaborative learning of multiple domain experts (ME) with the meta-learning strategy (ML). `EMA' means exponential moving average of domain experts to the universal expert. `$L_a$' and `$L_u$' denote the alignment and uniformity loss, respectively. `pro' denotes the projector. `UDCL' is the universal-domain collaborative learning, featured with EMA, $L_a$ and $L_u$. `M' is Market1501, `D' is DukeMTMC-reID, `MS' is MSMT17 and `C' is CUHK03. `MS+D+C$\to$M' means training on MSMT17, DukeMTMC-reID and CUHK03, and testing on Market1501.}
    \resizebox{0.99\textwidth}{!}{
      \begin{threeparttable}
        \begin{tabular}{l||l|cc|cc|cc|cc|cc}
            \hline
            \multirow{3}{*}{\textbf{No.}}&\multirow{3}{*}{\textbf{Experimental Settings}} &\multicolumn{8}{c|}{\textbf{Train/Test Domain Settings}}&\multicolumn{2}{c}{\multirow{2}{*}{\textbf{Average}}} \\ \cline{3-10}
            &&\multicolumn{2}{c|}{MS+D+C$\to$M}&\multicolumn{2}{c|}{MS+M+C$\to$D}&\multicolumn{2}{c|}{M+D+C$\to$MS}&\multicolumn{2}{c|}{MS+M+D$\to$C}&\\ \cline{3-12}
            &&~mAP~&Top-1 &~mAP~&Top-1&~mAP~&Top-1&~mAP~&Top-1&~mAP~&Top-1
            \\ \hline
            1&DAG Base.& 47.7&73.5& 45.7&63.5&8.7&22.9&29.3&29.3&32.9&47.3 \\
            2&MDE Base.& 53.9 &77.7& 52.0&67.7&12.7&31.4&28.9&29.6&36.9&51.6 \\
            3&MDE Base.+ML& 55.2&79.2&52.7&68.9&12.9&31.9&31.0&31.7&38.0&52.9 \\
            4$^*$&DDCL(ME+ML)&57.8&80.8&53.8&70.7&13.6&33.5&33.6&35.1&39.7&55.0\\
            5&DDCL+EMA& 58.7&81.5&54.4&71.4&15.1&36.2&35.6&36.8 &41.0&56.5\\
            6&DDCL w/o ML+EMA&37.0&62.7 & 41.6&58.1&8.2&21.9 &17.7&17.1&26.1&40.0\\
            7&DDCL+EMA+$L_a$ w/o pro& 59.1&82.0& 55.4&72.0&17.2&40.0&35.2&36.0&41.7&57.5 \\
            8&DDCL+EMA+$L_a$& 60.1&82.2&56.7&73.0&17.4&40.5&36.2&37.4&42.6&58.3 \\
            9&DDCL+UDCL(EMA+$L_a$+$L_u$)& \textbf{60.9}&\textbf{83.2}&\textbf{57.2}&\textbf{74.1}&\textbf{18.0}&\textbf{41.2}&\textbf{37.3}&\textbf{38.1} & \textbf{43.4}&\textbf{59.2}\\
            \hline
        \end{tabular}
            \begin{tablenotes}
        \item[*] Report the results of the best domain expert.
        \end{tablenotes}
        \end{threeparttable}
    }
    \label{tab:ablation study}
\end{table*}

\begin{table*}
    \centering
    \caption{Performance (\%) on the three domain experts. The results are permuted according to the source domains order. Take `MS+D+C' for example, the 1st domain expert belongs to MSMT17, and the 2nd and 3rd belongs to DukeMTMC and CUHK03. }
    \resizebox{0.99\textwidth}{!}{
        \begin{tabular}{l||l|cc|cc|cc|cc}
            \hline
            \multirow{2}{*}{\textbf{No.}}&\multicolumn{1}{c|}{\multirow{2}{*}{\textbf{Experimental Settings}}}&\multicolumn{2}{c|}{MS+D+C$\to$ M}&\multicolumn{2}{c|}{MS+M+C$\to$D}&\multicolumn{2}{c|}{M+D+C$\to$MS}&\multicolumn{2}{c}{MS+M+D$\to$C}\\ \cline{3-10}
            &&~mAP~&Top-1 &~mAP~&Top-1&~mAP~&Top-1&~mAP~&Top-1
            \\ \hline
            1&DDCL+EMA & 56.6/56.5/57.8& 81.1/80.0/80.8& 53.8/53.9/53.6& 70.7/71.1/71.1& 13.5/13.6/13.5& 33.0/33.5/33.1& 33.6/33.4/33.1&35.1/33.9/33.8 \\
            2&DDCL+EMA+$L_a$ w/o pro &58.9/58.6/59.2&82.2/82.1/81.9& 56.0/55.9/55.4&72.2/72.2/72.3&\textbf{16.6/16.6/16.5}& \textbf{38.9/38.7}/38.6 &35.1/35.1/35.4&35.4/\textbf{35.9/36.3}\\
            3&DDCL+UDCL(EMA+$L_a$) &\textbf{59.9/59.1/60.4}& \textbf{82.7/82.4/83.2}& \textbf{57.0/56.9/56.7}& \textbf{73.1/72.7/73.0}& 16.3/16.2/16.4& 38.7/\textbf{38.7/38.7}& \textbf{36.1/35.3/36.1}& \textbf{37.4/35.9}/36.1\\ 
            \hline
        \end{tabular}
    }
    \label{tab:domain-specific}
\end{table*}

\section{Experiments}
\subsection{Datasets \& Evaluation Metrics}
\label{sec:4.1}
\noindent \textbf{Datasets.} 
We follow the large-scale dataset setting of multi-source DG-ReID proposed in \citet{zhao2020learning}. This setting employs four large-scale person ReID datasets from different domains, including Market1501 \cite{market1501}, DukeMTMC-reID \cite{duke1, duke2}, CUHK03 \cite{CUHK03} and MSMT17 \cite{MSMT17}, which are widely used in recent ReID tasks. Following \citet{zhao2020learning, gulrajani2020search}, we use the leave-one-domain-out protocol to split the four datasets (domains) into training/testing domains, specifically, three datasets are used as source training domains and the left one is used as the unseen target domain. The detailed information of the large-scale dataset setting and another small-scale dataset setting \cite{song2019generalizable, lin2020multi}  in multi-source DG-ReID will be introduced in the Supplementary Materials.

\noindent \textbf{Evaluation Metrics.}
We follow the commonly used evaluation metrics in ReID to quantitatively evaluate the performance by mean Average Precision (mAP) and Cumulative Matching Characteristic (CMC) curve at Top-$k$.

\subsection{Implementation Details}
We utilize the ResNet50 \cite{he2016deep}, ResNet50-IBN \cite{pan2018two} and OSNet \cite{OSNet} as the backbones in the following experiments. The projector is a simple MLP network composed of Linear-BN-ReLU-Linear where the shapes of the two Linear layers are (2048, 512) and (512, 2048), respectively. We employ the iteration-based way to train those models where the max iteration is set to 12,000 per GPU and 8 GTX-1080TI GPUs are used. We optimize the model parameters of each domain-specific network by Adam \cite{kingma2014adam} optimizer with the weight decay $5 \times 10^{-4}$. The learning rate is initialized to $1 \times 10^{-5}$, and warmed up to $1 \times 10^{-3}$ gradually in the previous 1,000 iterations and then decay to $1 \times 10^{-4}$ and $1 \times 10^{-5}$ at the 4,000-th iteration and the 8,000-th iteration, respectively. At the beginning of each iteration, we randomly sample 32 images of 8 identities, i.e., 4 images per identity, from each source domain. Besides, some data augmentation methods used in conventional person ReID approaches are also employed, including random flipping, random cropping and random erasing \cite{zhong2020random}.

\subsection{Ablation Study}
We have conducted comprehensive ablation studies using ResNet50-IBN as the backbone to analyze each component of the MECL learning framework. Besides, This section also reflects how MECL come into being step by step. More ablation studies (e.g. visualization, hyper-parameter analysis) please refer to the Supplementary Materials.

\noindent \textbf{Comparison of Baseline Methods.} We firstly compare the two baseline methods, the traditional DAG baseline and our proposed MDE baseline.  As shown in Tab. \ref{tab:ablation study}-1, 2, our proposed MDE baseline outperforms the DAG baseline by a large margin on most current ReID benchmarks. On average, DAG falls behind MDE about 4.0\% and  4.3\% in mAP and Top-1 accuracy, indicating that MDE provides a stronger baseline than DAG in the multi-source DG-ReID task.

\noindent \textbf{Effectiveness of Meta-Learning.} Following MLDG \cite{DG_meta_li2018learning}, we also apply meta-learning \cite{finn2017model} strategy to train the MDE baseline to explore its effectiveness on domain generalization. As shown in Tab. \ref{tab:ablation study}-3, compared with the pure MDE baseline, training in the meta-learning manner obtains about 1.1\% in mAP and 1.3\% in Top-1 accuracy gains on average, proving that the meta-learning strategy is able to improve the model's generalization capability.

\begin{table*}
    \centering
    \caption{Compared with the state-of-the-arts on multi-source DG-ReID benchmarks. The first group are the results of SOTAs, the second and third are the results of MDE and MECL using different backbones.}
    \resizebox{0.95\textwidth}{!}{
      \begin{threeparttable}
        \begin{tabular}{l||l|cc|cc|cc|cc|cc}
            \hline
            \multirow{2}{*}{\textbf{Method}}&\multirow{2}{*}{\textbf{Backbone}} &\multicolumn{2}{c|}{MS+D+C$\to$ M}&\multicolumn{2}{c|}{MS+M+C$\to$D}&\multicolumn{2}{c|}{M+D+C$\to$MS}&\multicolumn{2}{c|}{MS+M+D$\to$C}& \multicolumn{2}{c}{\textbf{Average}}\\ \cline{3-12}
            &&~mAP~&Top-1 &~mAP~&Top-1&~mAP~&Top-1&~mAP~&Top-1&~mAP~&Top-1
            \\ \hline\hline
            QAConv&ResNet50&35.6&65.7&47.1&66.1&7.5&24.3&21.0&23.5&27.8&44.9\\
            CBN&ResNet50&47.3&74.7&50.1&70.0&15.4&37.0&25.7&25.2&34.6&51.7\\
            SNR&SNR& 48.5&75.2&48.3&66.7&13.8&35.1&29.0&29.1&34.9 & 51.5\\
            M$^3$L &ResNet50& 48.1 & 74.5 &50.5 & 69.4&12.9 &33.0& 29.9&30.7&35.4 & 51.9\\
            M$^3$L&ResNet50-IBN& 50.2& 75.9&51.1&69.2 &14.7& 36.9&32.1&33.1&37.0 & 53.8 \\
            OSNet&OSNet& 44.2&72.5&47.0&65.2&12.6&33.2&23.3&23.9& 31.8 & 48.7\\
            OSNet-IBN&OSNet-IBN& 44.9&73.0&45.7&64.6&16.2&39.8&25.4&25.7& 33.0 & 50.8\\
            OSNet-AIN&OSNet-AIN& 45.8&73.3&47.2&65.6&16.2&40.2&27.1&27.4& 34.1 & 51.6\\
            \hline\hline
            \multirow{4}{*}{MDE Base.}&ResNet50& 49.2&75.2&44.2&60.6&9.3&23.6&23.0&22.8&31.4&45.6\\
            &ResNet50-IBN& 53.9 &77.7& 52.0&67.7&12.7&31.4&28.9&29.6& 36.9 & 51.6\\
            &SNR& 53.8&77.7&52.5&69.5&16.8&39.5&30.5&30.4& 38.4 & 54.3\\
            &OSNet-IBN& 48.6&75.0&48.0&66.7&16.5&40.3&26.8&26.2& 35.0 & 52.0\\
            \hline\hline
             \multirow{4}{*}{MECL}&ResNet50&56.5&80.0&53.4&70.0&13.3&32.7&31.5&32.1& 38.7 & 53.7 \\
            &ResNet50-IBN& \textbf{60.9}&\textbf{83.2}&57.2&74.1&18.0&41.2&37.3&38.1 &43.4&59.2\\
            &SNR& 60.2&82.4&\textbf{57.6}&\textbf{75.0}&\textbf{21.7}&\textbf{47.7}&\textbf{38.3}&\textbf{38.5}& \textbf{44.5} & \textbf{60.9} \\
            &OSNet-IBN& 52.3&77.6&51.3&68.8&18.1&43.0&29.3&29.9& 37.8 & 54.8\\
            \hline
        \end{tabular}
            \begin{tablenotes}
        \item[]
        \end{tablenotes}
        \end{threeparttable}
    }
    \label{tab:sota}
\end{table*}

\begin{table}
    \centering
    \caption{Compared with SOTAs$^*$ on DG-classification benchmarks. \textbf{Top}: PACS; \textbf{Bottom:} OfficeHome.}
    \resizebox{0.95\linewidth}{!}{
      \begin{threeparttable}
        \begin{tabular}{l||ccccc}
            \hline
            \textbf{Method}&\textbf{Art}&\textbf{Cartoon}&\textbf{Photo}&\textbf{Sketch}&\textbf{Avg}
            \\ \hline
            ERM&83.2$\pm$1.3 &76.8$\pm$1.7&\textbf{97.2}$\pm$0.3&74.8$\pm$1.3&83.0 \\
            Mixup&85.2$\pm$1.9&77.0$\pm$1.7&96.8$\pm$0.8&73.9$\pm$1.6&83.2\\
            MLDG&81.4$\pm$3.6& 77.9$\pm$2.3 &96.2$\pm$0.3 &76.1$\pm$2.1& 82.9 \\
            MTL& 85.6$\pm$1.5&78.9$\pm$0.6& 97.1$\pm$0.3 &73.1$\pm$2.7 &83.7 \\
            RSC& 83.7$\pm$1.7&\textbf{82.9}$\pm$1.1&95.6$\pm$0.7&68.1$\pm$1.5&82.6 \\
            \hline
            MECL&\textbf{86.5}$\pm$1.2&80.5$\pm$0.8&96.2$\pm$0.3&\textbf{77.7}$\pm$0.1&\textbf{85.3} \\
            \hline\hline
            \textbf{Method}&\textbf{Art}&\textbf{Clipart}&\textbf{Product}&\textbf{RealWorld}&\textbf{Avg}
            \\ \hline
            ERM&61.1$\pm$0.9 &50.7$\pm$0.6 &74.6$\pm$0.3& 76.4$\pm$0.6& 65.7 \\
            Mixup& 61.4$\pm$0.5& 53.0$\pm$0.3& 75.8$\pm$0.2& 77.7$\pm$0.3& 67.0\\
            MLDG& 60.5$\pm$1.4& 51.9$\pm$0.2& 74.4$\pm$0.6& 77.6$\pm$0.4& 66.1 \\
            MTL&  59.1$\pm$0.3& 52.1$\pm$1.2& 74.7$\pm$0.4& 77.0$\pm$0.6& 65.7 \\
            RSC&  61.6$\pm$1.0& 51.1$\pm$0.8& 74.8$\pm$1.1& 75.7$\pm$0.9& 65.8 \\
            \hline
            MECL&\textbf{65.9}$\pm$1.3&\textbf{58.3}$\pm$0.5&\textbf{76.9}$\pm$1.0&\textbf{79.8}$\pm$0.4&\textbf{70.2} \\
            \hline
        \end{tabular}
        \begin{tablenotes}
        \item[*] All results of SOTAs are based on \citet{gulrajani2020search}.
        \end{tablenotes}
        \end{threeparttable}
    }
    \label{tab:sota-dg}
\end{table}

\noindent \textbf{Effectiveness of Domain-Domain Collaborative Learning.}  We argue that using a shared backbone can't learn each domain sufficiently, so we propose that each domain is associated with a specific network (domain expert), and employ the meta-learning strategy to strengthen the interaction among the multiple domain experts by the meta-test stage. Tab. \ref{tab:ablation study}-4 has reported the performance of the best domain expert, specifically, the performance increases by 1.7\% in mAP and 2.1\% in Top-1 accuracy on average with the help of the collaborative learning. Besides, meta-learning plays an important roles in the domain-domain collaborative learning, removing it causes great performance degradation as shown in Tab. \ref{tab:ablation study}-6.

\noindent \textbf{Effectiveness of EMA.} 
To study the effectiveness of UDCL, we firstly observe the performance of the universal expert after gathering the knowledge from the domain experts in the manner of exponential moving average (EMA) \cite{mean-teacher}. As shown in the Tab. \ref{tab:ablation study}-5, the universal expert outperforms the best domain experts by 1.3\% and 1.5\% in mAP and Top-1 accuracy on average, indicating that the universal expert is more generalizable than the domain experts for the reason that it has absorbed the knowledge of all domains.

\noindent \textbf{Effectiveness of Alignment Loss.} We have conducted two studies to show the powerful capability of the alignment loss which provides a supervision signal from the perspective of pushing positive samples. The first is to directly minimize the Euclidean distance between the features of the same image extracted from the universal expert and domain experts respectively. 
The second is that the features output from the domain experts will be transformed by a projector according to Eq. \ref{loss:align}. As shown in the Tab. \ref{tab:ablation study}-5,7,8, the alignment loss without projectors can bring slight improvements to most benchmarks , but it has a negative effect on CUHK03. However, with the addition of the projectors, the generalization capability of the universal expert is further improved, i.e., 1.6\% in mAP and 1.8\% Top-1 accuracy gains on average based on Tab. \ref{tab:ablation study}-5. Furthermore, we also observe the performance of the three domain experts as shown in the Tab. \ref{tab:domain-specific}. With the alignment loss, the generalization capability of the domain experts are improved as well. Notice that, the best domain experts in some experiments even outperform the universal expert when using the alignment loss. 

\noindent \textbf{Effectiveness of Uniformity Loss.} Different with the alignment loss, the uniformity loss supervises the training of the domain experts by fully exploiting the negative samples. As shown in Tab. \ref{tab:ablation study}-9, with the addition of the uniformity loss, the generalization capability of the universal expert are further improved. Specifically, the uniformity loss increases the performance by 0.8\% in mAP and 0.9\% in Top-1 accuracy on average.

\subsection{Compare with State-of-the-Arts}
To demonstrate the novelty and versatility of the proposed MECL, we compare it with some state-of-the-arts on both DG-ReID and DG-classification tasks.

\noindent \textbf{Results on Multi-Source DG-ReID.} We compare our proposed MECL with the state-of-the-arts (SOTAs) using multi-source DG-ReID setting, including QAConv \cite{QAConv},CBN \cite{CBN}, SNR \cite{SNR}, M$^3$L \cite{zhao2020learning} and OSNet \cite{OSNet}. Notice that, we adapt the single source based methods (e.g., CBN, SNR) to multi-source setting in the traditional DAG baseline manner. As shown in Tab. \ref{tab:sota}, MECL has achieved the best performance among these SOTAs under different types of backbones. Furthermore, because MECL is a model-agnostic training framework, some SOTAs with more generalizable networks can be trained using MECL to further improve the model's generalization capability, like SNR, OSNet, etc. Specifically, when SNR meets MECL, the performance is increased from 34.9\% to 44.5\% in mAP and from 51.5\% to 60.9\% in Top-1 accuracy on average. Besides, the proposed MDE baseline performs better on most of the backbones, indicating its superiority as the baseline of the multi-source DG-ReID task.

\noindent \textbf{Results on DG-Classification.} MECL is not ReID-specific, and it also can be applied to the traditional DG-classification task with slight modification, i.e., a classifier using EMA to update is added to the universal expert. The details of modification and training will be demonstrated in Supplementary Materials. The studies are conducted around two common benchmarks in the DG-classification task, PACS \cite{li2017deeper} and OfficeHome \cite{venkateswara2017deep}. As the results shown in Tab. \ref{tab:sota-dg}, MECL outperforms the current SOTAs by a large margin, including ERM \cite{gulrajani2020search}, Mixup \cite{mixup}, MLDG \cite{DG_meta_li2018learning}, MTL \cite{MTL} and RSC \cite{RSC}. Specifically, MECL respectively surpasses the best SOTAs (MTL \& Mixup) by 1.6\% and 3.2\% on the two benchmarks on average. 

\section{Conclusion}
In this paper, we have proposed a novel model-agnostic learning framework for multi-source DG-ReID, named \emph{\textbf{M}ultiple Domain \textbf{E}xperts \textbf{C}ollaboration \textbf{L}earning (MECL)}. Domain-Domain Collaborative Learning (DDCL) and Universal-Domain Collaborative Learning (UDCL) organically work together in MECL to improve the model's generalization capability. Extensive experiments on both DG-ReID and DG-classification benchmarks show that, without additional inference computation cost, our MECL framework significantly outperforms state-of-the-arts. We also establish a simple but rather strong multi-source DG-ReID baseline method named Multi-Domain Equality (MDE) that will facilitate future works in this area.

\bibliography{aaai22}

\end{document}